\documentclass[letterpaper, 10 pt, conference]{ieeeconf}  %

\IEEEoverridecommandlockouts                              %

\usepackage{graphics} %
\usepackage{epsfig} %
\usepackage{mathptmx} %
\usepackage{times} %
\usepackage{amsmath} %
\usepackage{amssymb}  %
\usepackage{url}

\title{\LARGE \bf
Object grasping planning for the situation when soft and rigid objects are mixed together
}

\author{Xiaoman Wang$^{1}$, Xin Jiang$^{1}$, \emph{Member, IEEE}, Jie Zhao$^{1}$, Shengfan Wang$^{1}$, and  Yunhui Liu$^{1,2}$, \emph{Fellow, IEEE}%
\thanks{$^{1}$Mechanical Engineering and Automation, Harbin Institute of Technology, Shenzhen 518055, China; {\tt\small 16b953039@stu.hit.edu.cn, x.jiang@ieee.org, \{zhaojie,18S053234\}@stu.hit.edu.cn}}%
\thanks{$^{2}$Department of Mechanical Engineering, The Chinese University of Hong Kong, Hong Kong, China;
        {\tt\small yhliu@mae.cuhk.edu.hk}}%
}

\begin{document}

\maketitle
\thispagestyle{empty}
\pagestyle{empty}

\begin{abstract}

In this paper, we propose a object detection method expressed as rotated  bounding box to solve grasping challenge in the scenes where rigid objects and soft objects are mixed together. Compared with traditional detection methods, this method can output the angle information of rotated objects and thus can guarantee that within each rotated bounding box, there is a single instance. This technology is especially useful in the case of pile of objects with different orientations. In our method, when uncategorized objects with specific geometry shapes (rectangle or cylinder) are detected, the program will conclude that some rigid objects are covered by the towels. If no covered objects are detected, the grasp planning is based on 3D point cloud obtained from the mapping between 2D object detection result and its corresponding 3D point cloud. Based on the information provided by the 3D bounding box covering the object, grasping strategy for multiple cluttered rigid objects, collision avoidance strategy are proposed. The proposed method is verified by the experiment in which rigid objects and towels are mixed together.

\end{abstract}

\section{INTRODUCTION}
In terms of logistics warehouse sorting, rubbish classifying, or household services, people are confronted with the environments where rigid objects and soft objects are mixed together. In this situation, people can grasp only rigid objects with ease and precision, and will not bring soft objects together. When people grasp a soft object, they can manage to avoid colliding surrounding rigid objects. By contrast, when a robot is facing the similar scene, it trends to grasp a rigid object with the surrounding soft object caught together, or collides with a rigid object when trying to grasp a soft object. The main reason involved in the fact is that for a robot it is difficult to distinguish a rigid object from the surrounding soft objects. In this paper, our proposal mainly contributes to the grasping planning problems in this condition.

\begin{figure}[tpb]
	\centering
	\includegraphics[width=8 cm]{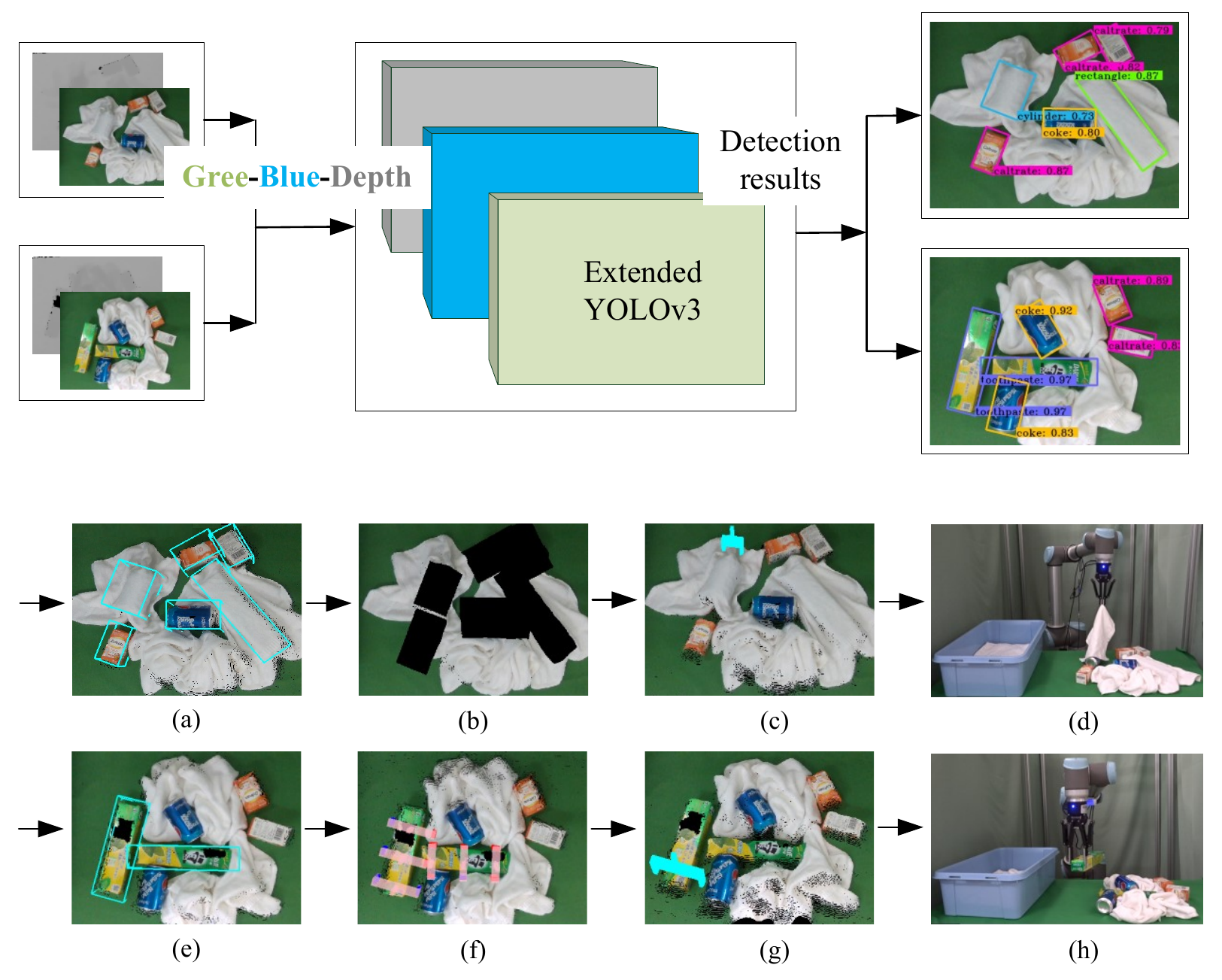}
	\caption{The overview of picking objects when the objects are composed of rigid and soft ones. (a) detection results of the unfeasible objects with 3D rotated bounding box; (b) filtering out the point cloud in 3D rotated bounding box; (c) grasping pose of towels in point cloud; (d) robot arm grasps a towel; (e)  detection results of the candidate grasping objects with 3D rotated bounding box; (f) grasping pose and collision detection of the candidate grasping objects; (g) appropriate grasping pose in point cloud; (h) robot arm grasps a rigid object.}
	\label{figurelabel00}
\end{figure} 

In our proposal, we make full use of object detection technology to filter out infeasible grasping plans (Fig. \ref{figurelabel00} demonstrates a typical grasping process under the scheme). At present, numerous object detection methods adopt bounding box to indicate the location of the detected targets in the image, such as YOLO \cite{redmon2016you,redmon2017yolo9000,redmon2018yolov3}, SSD  \cite{liu2016ssd}, etc. But the conventional bounding box is apt to cover multiple targets when they are randomly rotated. In addition, most of the above object detection methods use 2D image as the input of the network, when color of the object is similar with that of the background, the detection result is unsatisfied. Based on the rotated bounding box proposed by Lei, etc \cite{liu2017learning}, our method extends YOLOV3 \cite{redmon2018yolov3} and uses Green-Blue-Depth instead of RGB as the input of object detection network. With this method, it can detect each of the targets by covering its contour with a rotated bounding box even if they are overlapped with each other. 

The grasping planing methods for rigid and soft objects are generally different. There are many research work which contribute to solve the problem with only one type of targets are confronted \cite{tanwani2019fog, cai2019metagrasp, maitin2010cloth, chu2018real}. In this paper, we focus on the problem when both of two type objects are mixed together. In the situation shown in Fig. \ref{figurelabel00}, rigid objects may be covered with towels. If objects with specific regular geometric shape are detected on the surface of the towel, we can infer that there must be objects under the towel. Then grasping will be planned on the region excluding where rigid objects may exit. 
In this case, the grasping is targeted to the towel, which is solved by using our previous work \cite{wang2019picking}. If no targets with specific shape (rectangle or cylinder) are detected, the point cloud for determining candidate grasping target is obtained by using the object detection results, PCA is used to find the grasp poses of the targets. If no target is detected, which implies that there are only towels or nothing, our method determines where to grasp the towel.

In this work, we propose a method which outputs rotated bounding box as the result of object detection. We propose to utilize this technology to solve the grasping planning problem when rigid objects are mixed with towel. The main contributions of this paper are as follows:

\begin{itemize}

\item Our method represents multi-categories, multi-targets object detection result with rotated bounding box. The method uses Green-Blue-Depth instead of RGB as the input of network. 
\item The opening width of the gripper is determined by the size of the detected target, so it can effectively avoid the situation where multiple targets are grasped at one time.

\item Our method can avoid grasping multiple objects when rigid objects and soft objects are mixed.

\end{itemize}

\section{RELATED WORK}

At present, there are many studies on grasping planning, they mainly tackle the scenario of grasping an object from a pile of rigid objects or a bunch of soft objects. However, to the best of our knowledge, there is not a general method to address the mixed scene. In recent years, this problem, especially in the household services field, is becoming more and more important. We will briefly introduce some previous work in the field of grasping in the following.
 
\textbf{Grasping pose determination for rigid objects.} A. ten Pas, etc \cite{ten2017grasp} propose a method to determine grasp pose with point clouds, and this method does not use the CAD model. Based the work of \cite{ten2017grasp} and the network architecture of PointNet, H. Liang, etc \cite{liang2019pointnetgpd} proposed PointNetGPD, which can better understand the spatial geometry of the point cloud in the graspable region. J.Mahler, etc \cite{mahler2017dex} present a network GQ-CNN to obtain the grasping pose with the depth image and the method can obtain the grasping pose of the object with irregular shapes. There are many works for grasping pose detection evaluated on the Cornell grasping dataset \cite{web1}. These methods can represent grasping rectangle on the image \cite{redmon2015real} \cite{chu2018real} \cite{karaoguz2019object}. However, they are all for rigid objects.

\textbf{Grasping pose determination for soft objects.} At present, the research on grasping soft objects mainly is involved in the manipulation tasks of the clothes, such as folding, hanging, classification, etc. In these manipulation tasks, grasping usually plays an important role. Because clothes can be deformed, the determination of grasping pose is generally different from that of rigid objects. The selection of grasping pose of clothes depends on the purpose of grasping and the current state of clothes. The arbitrary point can be chosen as the grasping point as demonstrated in \cite{osawa2007unfolding}. However, this way may lead to air grasping or grasping with other unwanted object caught together. The center of the segmentation area are selected as the grasping point \cite{maitin2010cloth}. Instead of center point, the highest position of clothes \cite{bersch2011bimanual} or the point in the wrinkle \cite{wang2019picking} selected as the grasping point. During the folding process, the corner is usually used as the grasping point \cite{maitin2010cloth}. More detailed information on grasping clothes are introduced in survey \cite{jimenez2017visual}.

In practical scenarios, such as logistics sorting, garbage sorting, household services, there exit scenes where rigid and soft objects are mixed together. Obviously, the method mentioned above can not realize the sorting of rigid and soft objects. In the above scenario, the opening width of the gripper needs to be changed with the variation of the object. For example, when grasping towels, the opening width of the gripper should not be too large, otherwise, multiple towels will be grasped. When grasping rigid objects, it is necessary to ensure that the opening range of the gripper is larger enough than the minimum width of the object. Therefore, in this paper, we propose a method using the object detection expressed as rotated bounding box to solve the grasping problem that rigid and soft objects are mixed together.

\section{METHOD}
\subsection{Problem Statement}

\begin{figure}[tpb]
	\centering
	\includegraphics[width=8 cm]{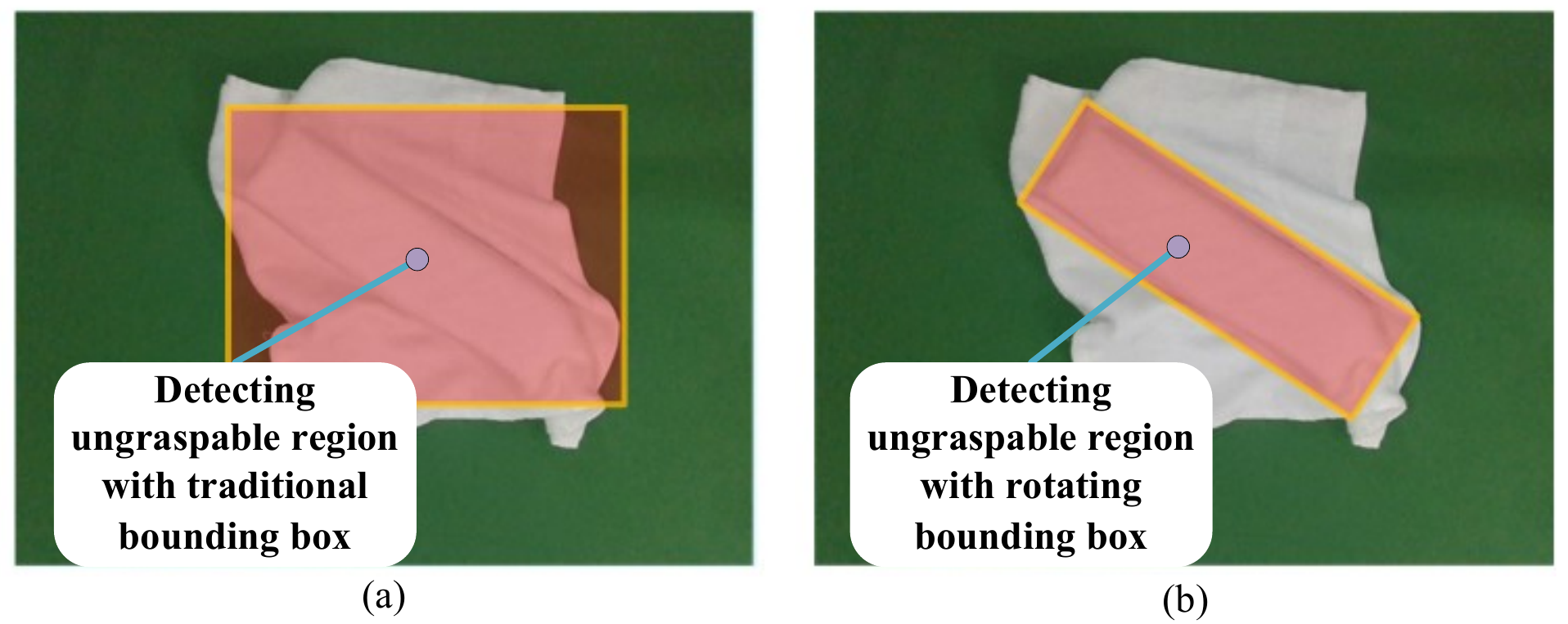}
	\caption{The difference between our method and previous work in detecting ungraspable region}
	\label{figurelabel01}
\end{figure}

In the scenario where rigid objects are mixed with towels, our method aims to grasp the towel without colliding to the occluded objects or grasp a rigid object without bringing soft object together. If objects are covered with towels, it can cause collision with occluded object when grasping the towel. So it is important to identify the area where occluded objects exist. A lot of object detection methods adopt bounding box to represent the location of the objects. As mentioned in \cite{liu2017learning}, a traditional bounding box has four variables: the center point($x,y$), and the size of the box($w,h$). It can not provide an accurate location of the object when it is rotated(see Fig \ref{figurelabel01}(a)). So the accessible grasping region becomes very small(see Fig \ref{figurelabel01}(a)). However, a rotated bounding box can solve this problem, as shown in Fig \ref{figurelabel01}(b).

\subsection{Rotated Bounding Box Prediction and Angle-related IOU}

\begin{figure}[tpb]
	\centering
	\includegraphics[width=7 cm]{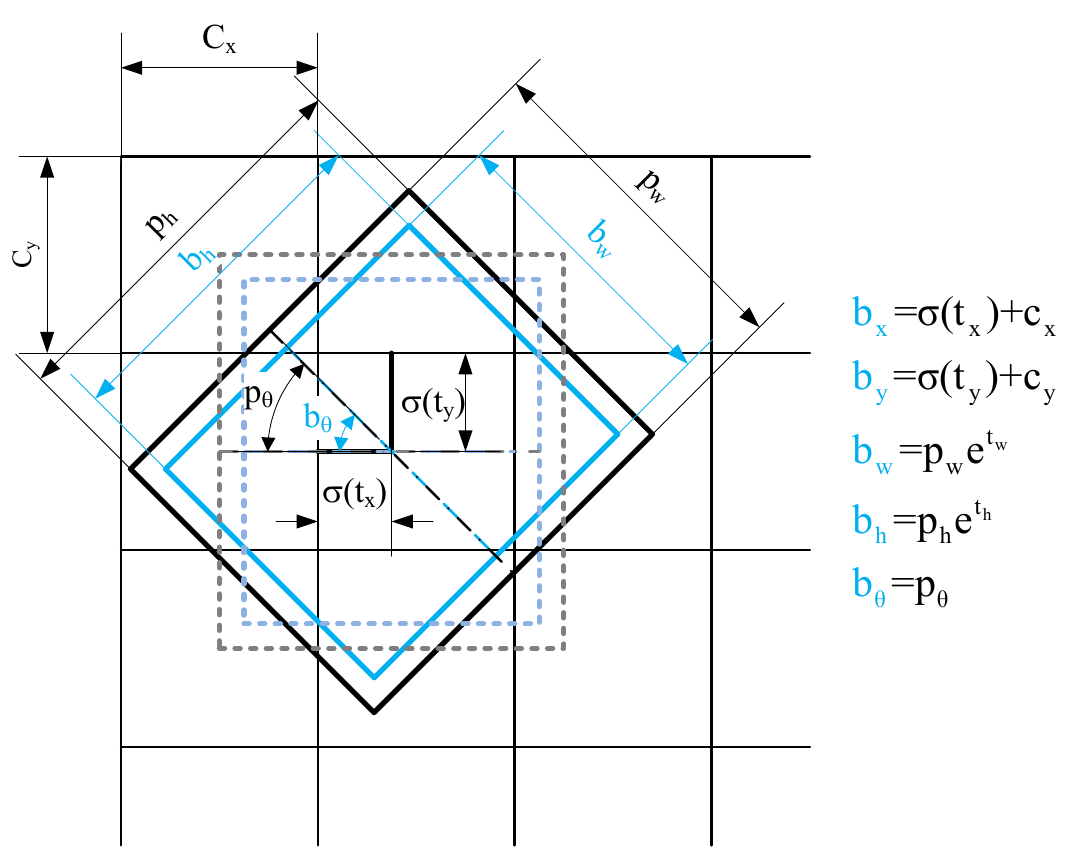}
	\caption{The rotated bounding boxes with dimension priors and location. The prediction of width, height and coordinates is exactly the same as of YOLOv3 \cite{redmon2018yolov3}. The difference is that the input of the network is Green-Blue-Depth, the training label has one more angle information, and the calculation of IOU is replaced by  Angle-related IOU, and the predicted boxes are rotated bounding box.}
	\label{figurelabel02}
\end{figure}

The rotated bounding box (this inspiration comes mainly from \cite{liu2017learning}) has five variables: the center point($b_x,b_y$), the size of the box($b_w,b_h$), and the angle $b_\theta$ of the bounding box(see Fig \ref{figurelabel02}). The training of object detection with rotated bounding box is extended from YOLOv3 \cite{redmon2018yolov3} training procedure by introducing angle estimation. In YOLOv3 \cite{redmon2018yolov3}, anchor boxes determined by dimension clusters are used to predict bounding boxes, we also use the same way to determine anchor boxes. However, we add an angle to each anchor box, that is,the anchor box has 5 parameters($a_{c_x},a_{c_y},a_w,a_h,a_\theta$). The angles of the 9 anchor boxes are 10,30,50,70,90,110,130,150,170 degrees, respectively.

The extended network predicts 5 parameters, 4 coordinates($t_x,t_y,t_w,t_h$) and a rotation angle $\theta$ for each bounding box. As mentioned in YOLOv3 \cite{redmon2018yolov3}, if the coordinates of the cell deviating from the top corner of the image are ($c_x,c_y$), and the size and the rotation angle of the rotated bounding box prior are ($p_w,p_h$) and $p_\theta$, then the predicted result is:

\begin{eqnarray}
b_{x} = \sigma \left ( t_{x} \right ) + c_{x}\\
b_{y} = \sigma \left ( t_{y} \right ) + c_{y}\\
b_{w} =p_{w} e^{t_{w}}\\
b_{h} =p_{h} e^{t_{h}}\\
b_{\theta} =p_{\theta} 
\label{eqnarray1}
\end{eqnarray}

During training we also use sum of squared error loss as the same as YOLOv3. In object detection, the accuracy of object detection is usually measured by IOU. The higher the IOU, the more precise the boundary box is. Because the rotated bounding box is angle-related, the traditional definition of IOU is no longer applicable. Angle-related IOU(ArIOU) is defined in the literature \cite{liu2017learning}. It is shown below:

\begin{eqnarray}
{\rm ArIOU} \left ( A,B \right ) = \frac{area\left ( A\cap B \right )}{area\left ( A\cup  B \right )}\left | \cos\left ( \theta _{A}-\theta _{B} \right ) \right |
\label{eqnarray2},
\end{eqnarray} where $\theta _{A}$ and $\theta _{B}$ are the rotated angle of the rotated bounding box A and B. If the estimated bounding box and the actual one overlap perfectly, then the ArIOU is 1. If the angle between estimated bounding box and the actual one is 90 degrees or there is no intersection between the two bounding boxes, then the ArIOU is 0.

\subsection{Grasping Pose}

\textbf{Grasping pose of towels}. Our method is to find the appropriate grasping pose on the wrinkles based on point cloud. Selection of the candidate wrinkle point cloud for towels can be referred to our previous research \cite{wang2019picking}. In this paper, PCA  is used to obtain the grasping points of candidate wrinkle instead of choosing the most convex point as in our previous research  \cite{wang2019picking}. Because the most convex points may be located at the edge of the wrinkle, its position is not a proper grasping position. In contrast, the grasp point obtained by PCA is generally located in the center of the wrinkle, which is easier to grasp. In the work \cite{wang2019picking}, the grasping direction of two-fingered gripper is determined by the normal vectors ($n_x,n_y$) of the grasping point. But this method does not guarantee that the opening direction of the two-fingered gripper is always perpendicular to the wrinkle. In this paper, PCA is also used to find the rotation direction of candidate wrinkle point cloud, and the rotated angle of the wrinkle is taken as the rotated angle of two-fingered gripper. This method can effectively avoid failure due to incorrect rotation angle of two-fingered gripper.

\textbf{Grasping pose of rigid objects}. When towels and other items are mixed together, the selection of grasping pose for rigid objects is different from that for soft objects. Soft objects can be deformed and adapted to grippers, but these characteristics are not applicable for rigid objects. Our method first detects the object's rotated boundary box, and then takes the object with the highest score as the candidate target for grasping. Then the 2D position information of the target is mapped to the 3D point cloud, and 3D rotated boundary box of the target also helps to isolate it from the background. Finally, the point cloud of the candidate grasping target expressed as 3D rotated boundary box is extracted and the grasping point and orientation is determined by PCA. The size $(w_r,h_r)$ of target is determined by the size of 3D rotated bounding box. It should be noted that the grasping point is the central point $P_c$ ($p_{c_x},p_{c_y},p_{c_z}$), and the grasping direction ($n_x,n_y$) is primary direction of the extracted point cloud. But this grasp point is not always a proper one, especially when two objects are close to each other, it is apt to lead to collision. So we sample three grasp positions($P_1$ ($p_{c_x},p_{c_y},p_{c_z}$), $P_2$ ($p_{c_x} - 0.3*h_r,p_{c_y}-0.3*h_r,p_{c_z}-0.3*h_r$), $P_3$ ($p_{c_x} + 0.3*h_r,p_{c_y}+0.3*h_r,p_{c_z}+0.3*h_r$)) with equal spacing in the primary direction of extracted point cloud to detect whether there exist collision. If there exist collisions for all the three positions, our method chooses another object with lower detection scores as target for grasping.

\textbf{Collision avoidance of grasping pose}. Because the soft object is adaptive to the gripper, collision avoidance is not required. In this paper, collision detection is only applied to rigid objects. Let $V\left ( L \right )\subset R^3$, $V\left ( R \right )\subset R^3$ represent the the volume occupied by the left and right finger of two-fingered gripper, and $V\left ( L \right )\subset R^3$, $V\left ( R \right )\subset R^3$ are all 3D rotated boundary boxes. Let $N\subset R$ represents the number of point cloud in the 3D rotated boundary box. If ($N\left ( L \right ) \leq C_T$ and $N\left ( R \right ) \leq C_T$), where $C_T$ is a constant and the value is 60, the grasping pose is collision free. Because there is noise in the point cloud, a constant $C_T$ is set to increase the tolerance of collision avoidance.

\section{EXPERIMENTAL RESULTS}
\subsection{Detection results}
\begin{figure}[tpb]
	\centering
	\includegraphics[width=8 cm]{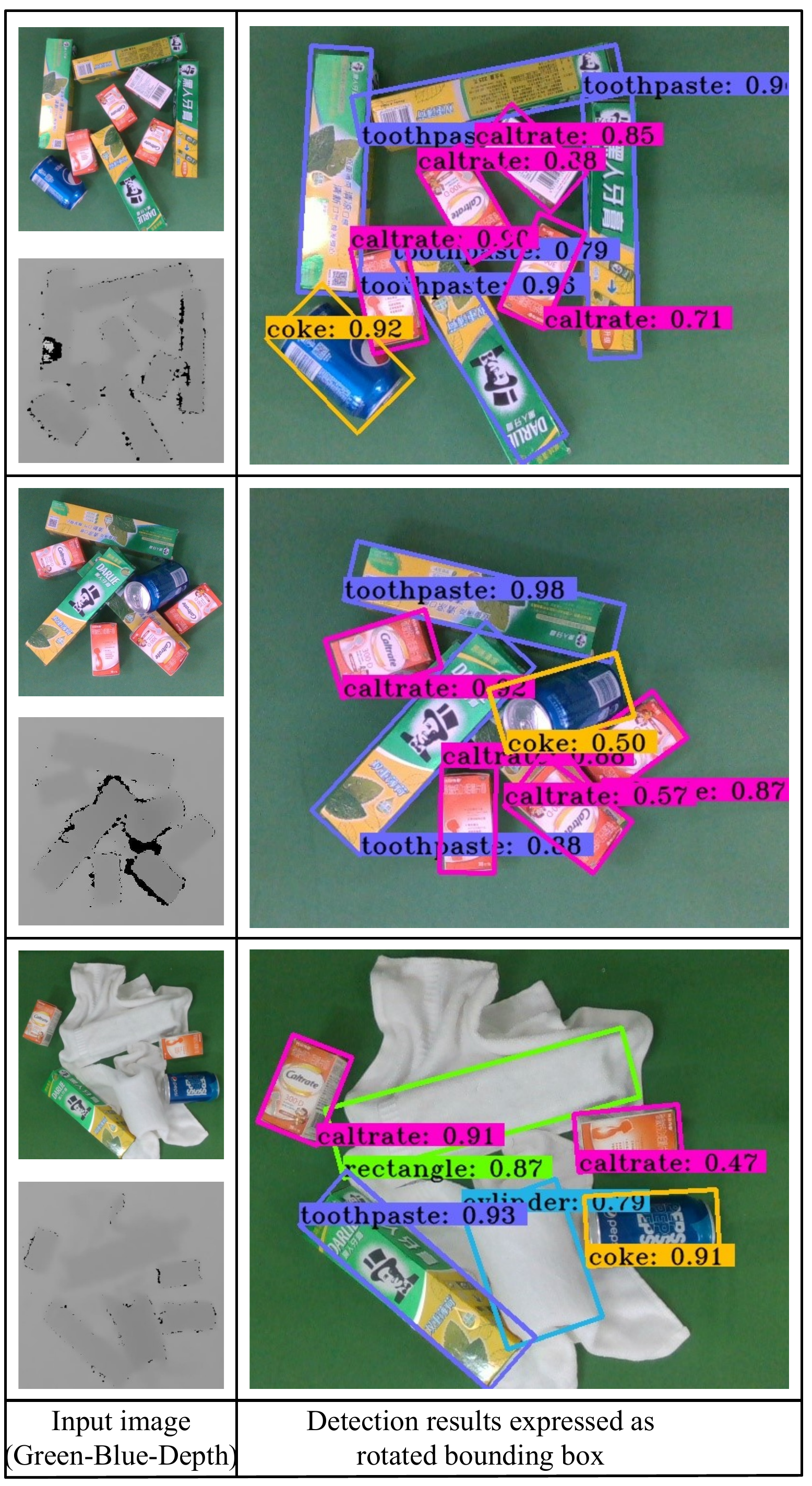}
	\caption{The result of proposed object detection method using rotated bounding box.}
	\label{figurelabel03}
\end{figure}

 As can be seen from Fig \ref{figurelabel01}, both the color of the occluded objects and the background may be same, that is, all of them are white. Therefore, if only RGB is used as input data of the network, the object detection algorithm will not distinguish between them. To use the YOLOv3 with RGB-D data we only need to replace the red channel in the image with the depth information, and the idea comes mainly from \cite{redmon2015real}. Although we can also modify the architecture to support RGB-D with four channels, but then we can not use pre-trained weights. The detection results with rotated bounding box are shown in the Fig \ref{figurelabel03}.
 
\subsection{Towels grasping}

\begin{figure}[tpb]
	\centering
	\includegraphics[width=8 cm]{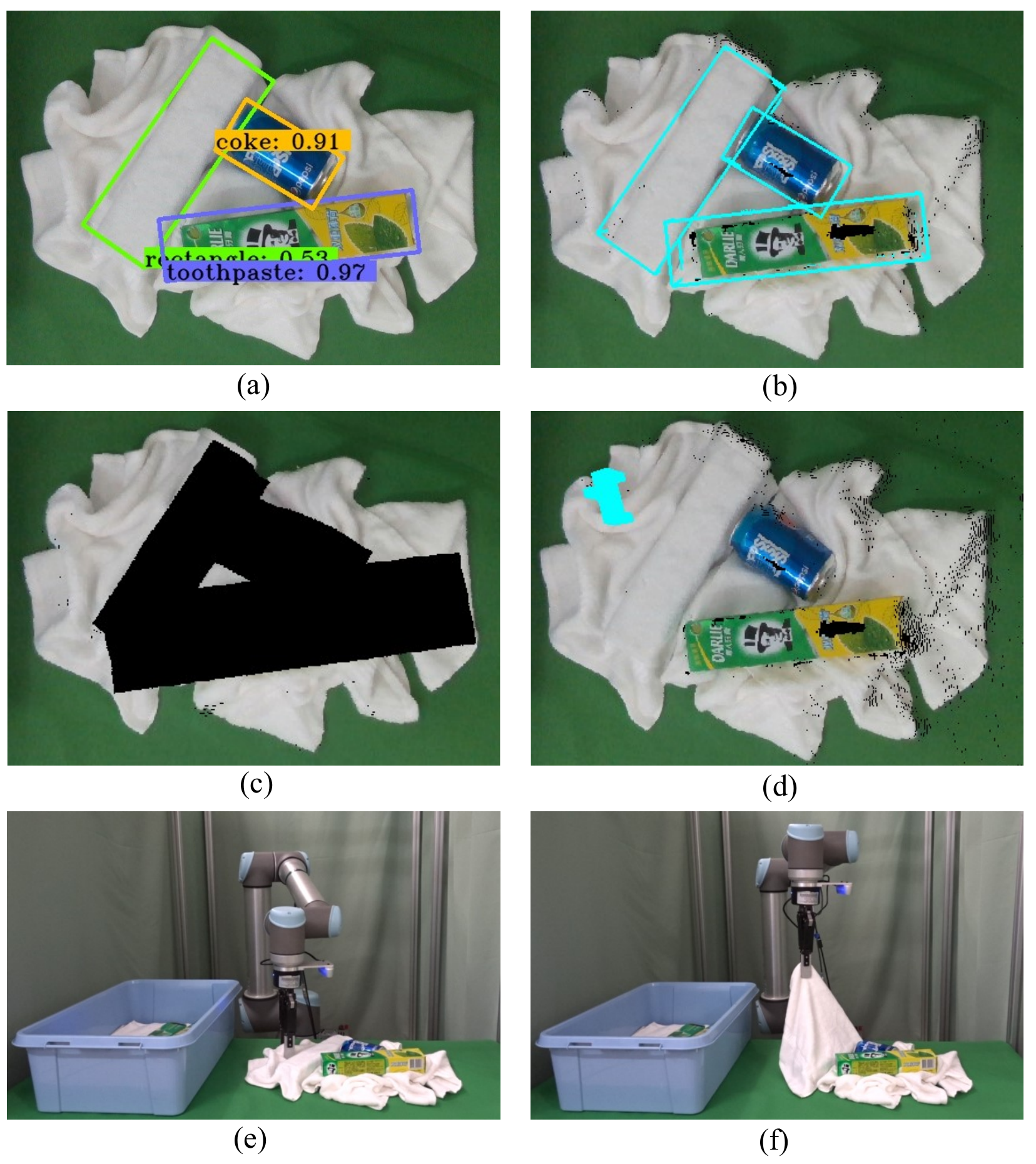}
	\caption{The overview of grasping a towel in this scenario where the rigid objects are all covered with towels. (a) detection results using rotated bounding box; (b) detection results using 3D rotated bounding box; (c) filtering out the point cloud in 3D rotated bounding box; (d) grasping pose in point cloud;
		(e) robot arm moves to the grasping pose; (f) robot arm grasps a towel.}
	\label{figurelabel04}
\end{figure}

When the multiple rigid objects are covered with towels, in order to avoid colliding to the occluded objects, the position of the occluded objects are needed to be detected. However, the traditional bounding box is not accurate in locating objects. Therefore, we replaced it with the rotated bounding box and the detection results is shown in Fig \ref{figurelabel04}(a).

 If the detected objects includes a rectangle or cylinder, it indicates that there are objects covered by towels. It is only necessary to exclude the region marked as unfeasible for grasping, that is, the region where the specific geometries (rectangle or cylinder) and rigid objects are detected. Let $V{}\left (sob \right )\subset R^3$ denote the rotated boundary boxes of the specific objects (rectangle or cylinder). Let $V\left (rob \right )\subset R^3$ denote the rotated boundary boxes of the rigid objects (such as toothpaste, coke, etc). Let $V{}'\left (rob \right )=1.5\ast1.5\ast$V$\left (rob \right )$, that is, expanding ($w_r,h_r$) of $V\left (rob \right )$ by 1.5 times. Let $C\left ( sob \right )\subset R^3$ denote the point cloud in $V\left (sob \right )$. Let $C\left ( rob \right )\subset R^3$ denote the point cloud in $V\left (rob \right )$. Because the rotation angle detected is not very accurate, so the point cloud $C\left ( rob \right )$ may not be consistent with the object. However, residual point cloud may affect determining grasping pose of the towel. Let $C{}'\left ( rob \right )\subset R^3$ denote the point cloud in $V{}'\left (rob \right )$, which replace $C\left ( rob \right )$  as the point cloud of the rigid objects. So the feasible region for grasping in point cloud can be represented as: $C\left ( g \right ) = C \cap C(sob)\cap C{}'(rob)$(see Fig \ref{figurelabel04}(c)). The whole process of grasping a towel is shown in Fig \ref{figurelabel04}. In addition, the opening width of the gripper is set as a little larger than the width of the wrinkle(in our test, generally gripper opening width:$g_w =30$ mm). The towel is deformable and adaptive to the gripper, so collision detection is not required.

\subsection{Rigid objects grasping}

Rigid objects are non-deformable and have no adaptability to gripper. Therefore, for multiple objects placed together, if the opening width of the gripper and grasping pose are not suitable, it may cause multiple objects to be grasped together or fail to grasp.

\begin{figure}[tpb]
	\centering
	\includegraphics[width=8 cm]{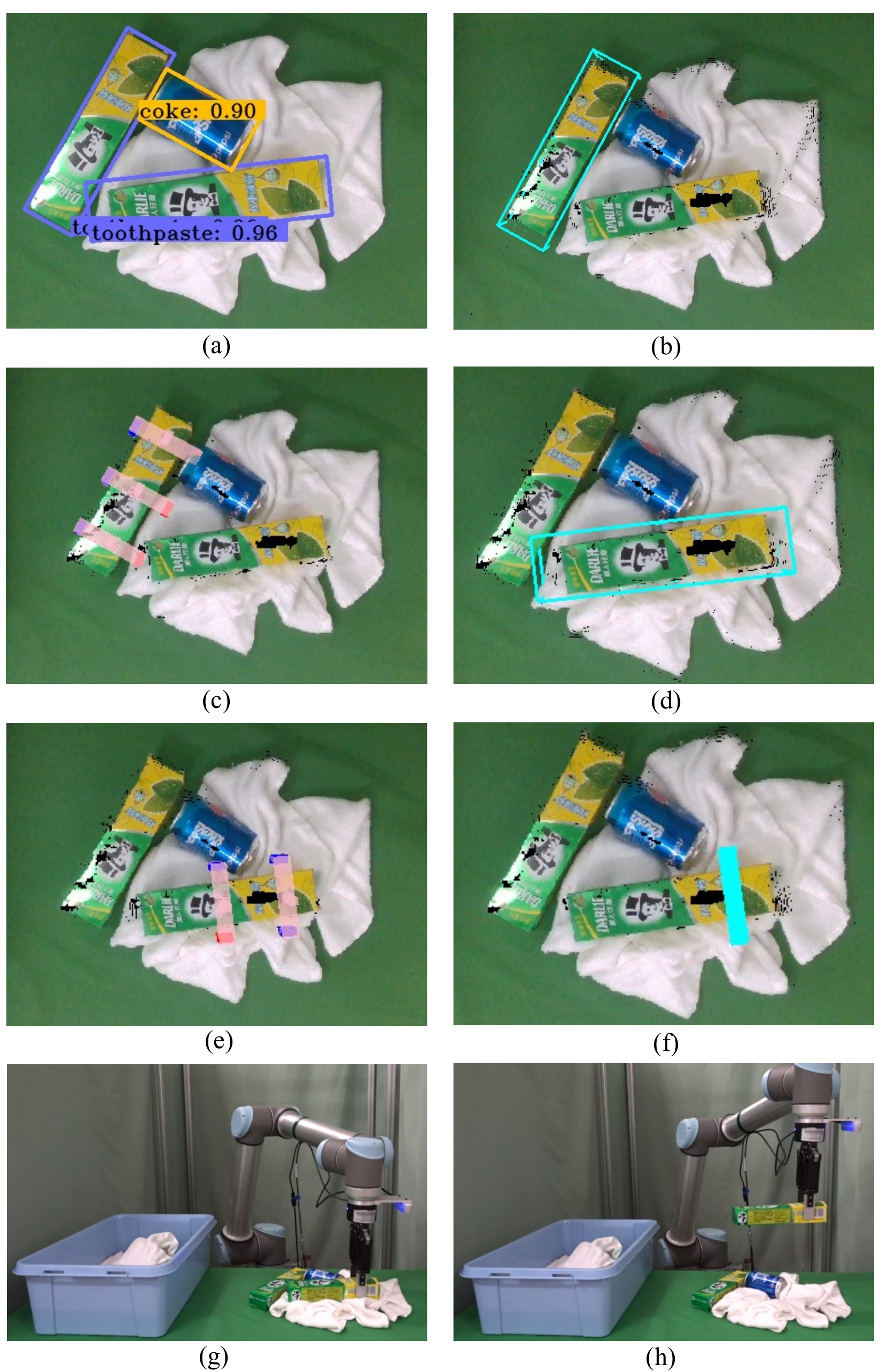}
	\caption{The whole process of grasping a object. (a) detection results using rotated bounding box; (b) the first candidate grasping object in 3D rotated bounding box; (c) grasping pose and collision detection of the first candidate grasping object; (d) the second candidate grasping object in 3D rotated bounding box; (e) grasping pose  and collision detection of the second candidate grasping object; (f) appropriate grasping pose in point cloud; (g) robot arm move to the grasping pose; (h) robot arm grasps a object.}
	\label{figurelabel05}
\end{figure}

In this paper when grasping a rigid object, the opening width of the gripper is determined by the size ($w_r,h_r$) of rotated bounding box, it is 1.5 times width ($1.5*w_r$) of the object. In addition, for a rigid object, the appropriate grasping pose should be collision free ones. When the detected objects are not including rectangle or cylinder, the object with the highest score (toothpaste, see Fig \ref{figurelabel05}(a)) is selected as the candidate grasping target. And then the point cloud of candidate grasping target in 3D rotated bounding box can be obtained, see Fig \ref{figurelabel05}(b)). The grasping pose can be determined using the extracted point cloud by the PCA, see Fig \ref{figurelabel05}(c)). If a collision is detected, the finger will become to red, otherwise the finger will be blue. As Shown in Fig \ref{figurelabel05}(c)), the grasping poses are all collisional ones. Therefore, the next candidate grasping target with the second highest score need to be sampled (see Fig \ref{figurelabel05}(d), (e)). Since the grasping pose of the second candidate grasping target is collision free, so it is chosen as the final grasping pose(see Fig \ref{figurelabel05}(f)).

\begin{figure}[tpb]
	\centering
	\includegraphics[width=8 cm]{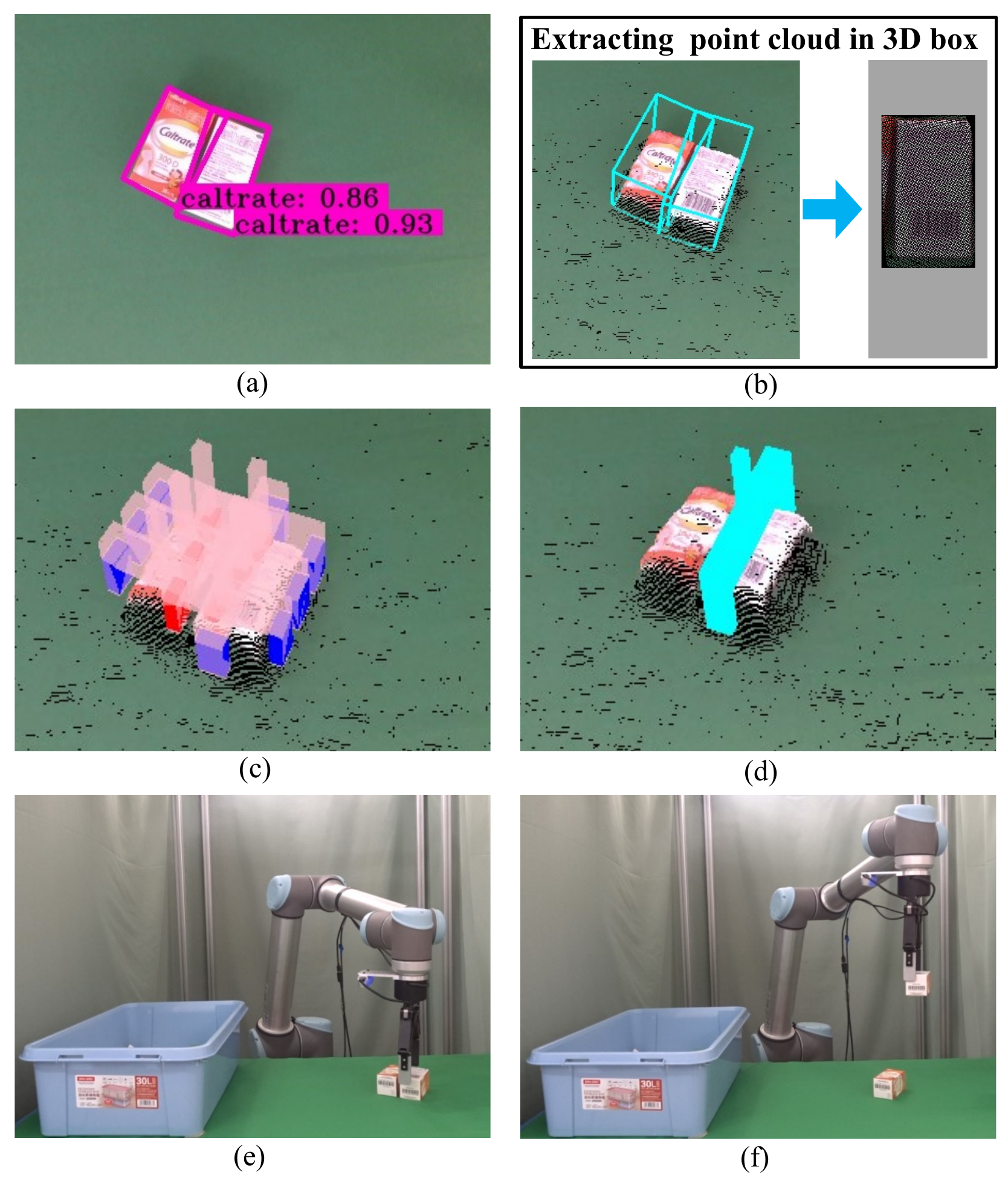}
	\caption{The whole process of grasping a object. (a) detection results using rotated bounding box; (b) detection results using 3D rotated bounding box and extracting point cloud in 3D rotated bounding box; (c) collision detection of the grasping pose; (d) grasping pose in point cloud; (e) robot arm moves to the grasping pose; (f) robot arm grasps a object.}
	\label{figurelabel06}
\end{figure}

\begin{figure}[tpb]
	\centering
	\includegraphics[width=8 cm]{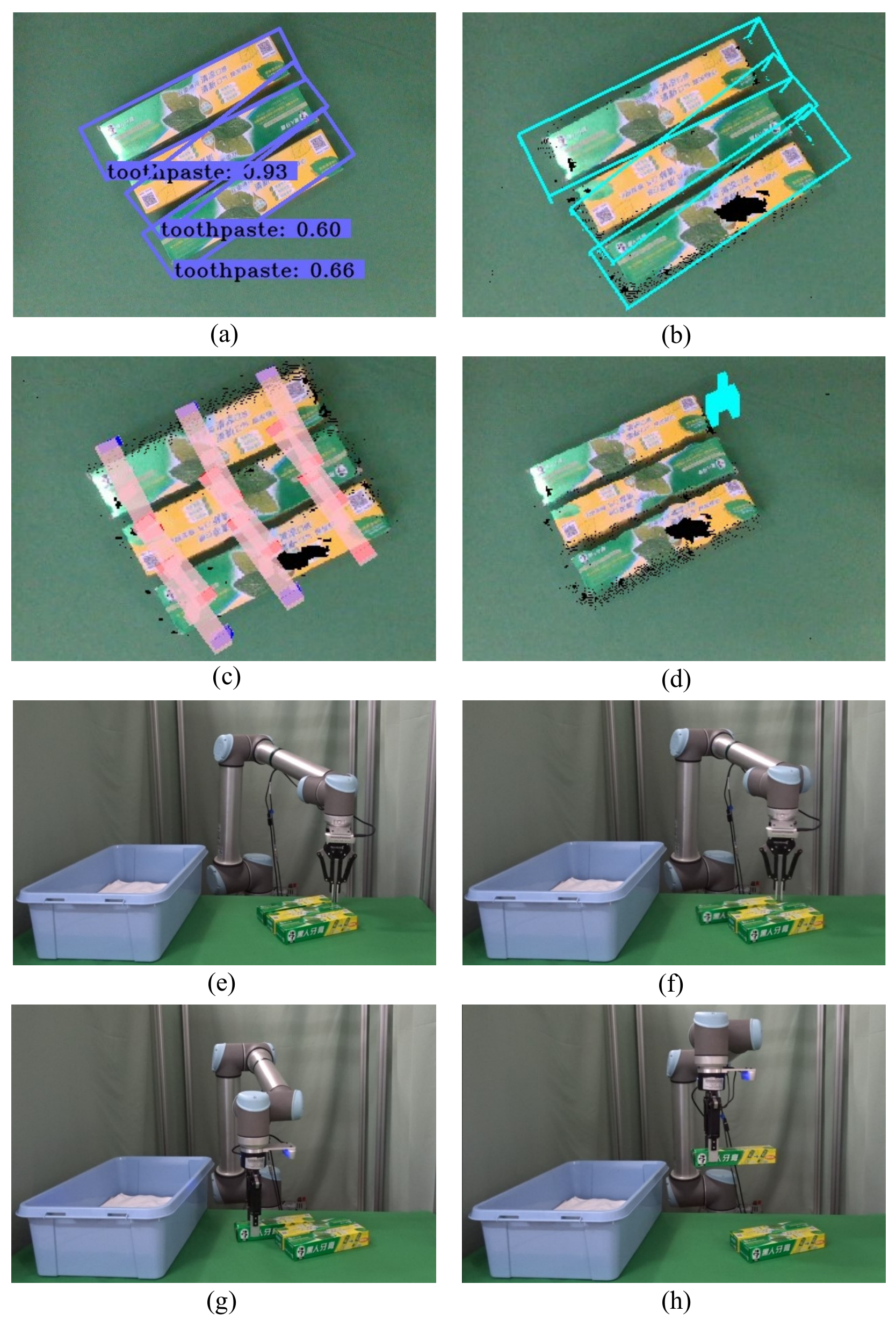}
	\caption{The whole process of grasping a object. (a) detection results using rotating bounding box; (b) detection results using 3D rotating bounding box and extracting point cloud in 3D rotating bounding box; (c) collision detection of the first grasping pose; (d) collision detection of the second grasping pose; (e) collision detection of the third grasping pose; (f) grasping pose in point cloud; (g) robot arm moves to the grasping pose; (h) robot arm grasps a object.}
	\label{figurelabel07}
\end{figure}

In the process of grasping rigid objects, if two or more objects are close too to each other, it is difficult to grasp them limited by gripper in this case(see Fig \ref{figurelabel06}(a) and Fig \ref{figurelabel07}(a)). There are two solutions for this scenario. 

\textbf{The first solution.} If detecting collision for every object in three time(see Fig\ref{figurelabel06}(c)), there is still no proper grasping position. In this case, the gripper width should be change as 1.2 times length ($1.2*h_r$) of the object, and the grasping point and the grasping direction are still determined using the extracted point cloud , the direction should be rotated by 90 degrees(see Fig\ref{figurelabel06}(d)). Except for detecting collision, it is necessary to detect whether the opening width of the gripper exceeds the opening limit. The whole grasping process is shown in Fig \ref{figurelabel06}.

\textbf{The second solution.} There are some occasions when all the above methods had been tried, it still fail to find a appropriate grasping pose (see Fig\ref{figurelabel07}(c)). In this case, pushing and grasping strategy is adopted in our work. The object with the highest score is regarded as the push target. The push direction ($n_x,n_y$) and the push ending pose P\_end ($P_{e_x} =p_{c_x}, P_{e_y} =p_{c_y}, P_{e_z} = p_{c_z}$) can be determined by computing PCA of the push target point cloud in 3D rotated bounding box. The starting pose of pushing is P\_start ($P_{s_x} =p_{c_x} + n_x *0.5*h_r, P_{s_y} =p_{c_y} + n_y * 0.5*h_r, P_{s_z} = pc_z$).

\section{CONCLUSIONS}

In this paper, we propose a object detection method to solve the grasp planning problem when soft and rigid objects are mixed together.  The object detection method is the extension of YOLOv3, which can detect objects covered by towels via rotated bounding box. When grasping towels, our method can eliminates the unfeasible grasping region, due to the potential collision to the covered objects. If no covered objects are detected, the grasping poses of candidate rigid objects can be determined by mapping 2D object detection result to its corresponding 3D point cloud. Beyond that, the opening width of the gripper varies with the size of the grasping target, which can effectively avoid grasping multiple objects.

\addtolength{\textheight}{-12cm}   %

\section*{ACKNOWLEDGMENT}

This research was funded by a Shenzhen Peacock Plan Team grant (KQTD20140630150243062), the Shenzhen and Hong Kong Joint Innovation Project (SGLH20161209145252406), and the Shenzhen Fundamental Research grant (JCYJ20170811155308088).

\bibliographystyle{IEEEtran}
\bibliography{reference}

\begin{thebibliography}{10}
\providecommand{\url}[1]{#1}
\csname url@rmstyle\endcsname
\providecommand{\newblock}{\relax}
\providecommand{\bibinfo}[2]{#2}
\providecommand\BIBentrySTDinterwordspacing{\spaceskip=0pt\relax}
\providecommand\BIBentryALTinterwordstretchfactor{4}
\providecommand\BIBentryALTinterwordspacing{\spaceskip=\fontdimen2\font plus
\BIBentryALTinterwordstretchfactor\fontdimen3\font minus
  \fontdimen4\font\relax}
\providecommand\BIBforeignlanguage[2]{{%
\expandafter\ifx\csname l@#1\endcsname\relax
\typeout{** WARNING: IEEEtran.bst: No hyphenation pattern has been}%
\typeout{** loaded for the language `#1'. Using the pattern for}%
\typeout{** the default language instead.}%
\else
\language=\csname l@#1\endcsname
\fi
#2}}

\bibitem{redmon2016you}
J.~Redmon, S.~Divvala, R.~Girshick, and A.~Farhadi, ``You only look once:
  Unified, real-time object detection,'' in \emph{Proceedings of the IEEE
  conference on computer vision and pattern recognition}, 2016, pp. 779--788.

\bibitem{redmon2017yolo9000}
J.~Redmon and A.~Farhadi, ``Yolo9000: better, faster, stronger,'' in
  \emph{Proceedings of the IEEE conference on computer vision and pattern
  recognition}, 2017, pp. 7263--7271.

\bibitem{redmon2018yolov3}
------, ``Yolov3: An incremental improvement,'' \emph{arXiv preprint
  arXiv:1804.02767}, 2018.

\bibitem{liu2016ssd}
W.~Liu, D.~Anguelov, D.~Erhan, C.~Szegedy, S.~Reed, C.-Y. Fu, and A.~C. Berg,
  ``Ssd: Single shot multibox detector,'' in \emph{European conference on
  computer vision}.\hskip 1em plus 0.5em minus 0.4em\relax Springer, 2016, pp.
  21--37.

\bibitem{liu2017learning}
L.~Liu, Z.~Pan, and B.~Lei, ``Learning a rotation invariant detector with
  rotatable bounding box,'' \emph{arXiv preprint arXiv:1711.09405}, 2017.

\bibitem{tanwani2019fog}
A.~K. Tanwani, N.~Mor, J.~Kubiatowicz, J.~E. Gonzalez, and K.~Goldberg, ``A fog
  robotics approach to deep robot learning: Application to object recognition
  and grasp planning in surface decluttering,'' \emph{arXiv preprint
  arXiv:1903.09589}, 2019.

\bibitem{cai2019metagrasp}
J.~Cai, H.~Cheng, Z.~Zhang, and J.~Su, ``Metagrasp: Data efficient grasping by
  affordance interpreter network,'' \emph{arXiv preprint arXiv:1902.06554},
  2019.

\bibitem{maitin2010cloth}
J.~Maitin-Shepard, M.~Cusumano-Towner, J.~Lei, and P.~Abbeel, ``Cloth grasp
  point detection based on multiple-view geometric cues with application to
  robotic towel folding,'' in \emph{2010 IEEE International Conference on
  Robotics and Automation}.\hskip 1em plus 0.5em minus 0.4em\relax IEEE, 2010,
  pp. 2308--2315.

\bibitem{chu2018real}
F.-J. Chu, R.~Xu, and P.~A. Vela, ``Real-world multiobject, multigrasp
  detection,'' \emph{IEEE Robotics and Automation Letters}, vol.~3, no.~4, pp.
  3355--3362, 2018.

\bibitem{wang2019picking}
X.~Wang, X.~Jiang, J.~Zhao, S.~Wang, T.~Yang, and Y.~Liu, ``Picking towels in
  point clouds,'' \emph{Sensors}, vol.~19, no.~3, p. 713, 2019.

\bibitem{ten2017grasp}
A.~ten Pas, M.~Gualtieri, K.~Saenko, and R.~Platt, ``Grasp pose detection in
  point clouds,'' \emph{The International Journal of Robotics Research},
  vol.~36, no. 13-14, pp. 1455--1473, 2017.

\bibitem{Liang2018PointNetGPD}
H.~Liang, X.~Ma, S.~Li, M.~Go¨Rner, and S.~Tang, ``Pointnetgpd: Detecting
  grasp configurations from point sets,'' 2018.

\bibitem{mahler2017dex}
J.~Mahler, J.~Liang, S.~Niyaz, M.~Laskey, R.~Doan, X.~Liu, J.~A. Ojea, and
  K.~Goldberg, ``Dex-net 2.0: Deep learning to plan robust grasps with
  synthetic point clouds and analytic grasp metrics,'' \emph{arXiv preprint
  arXiv:1703.09312}, 2017.

\bibitem{web1}
``Cornell grasping dataset.''
  \url{http://pr.cs.cornell.edu/grasping/rect_data/data.php},
  accessed:2013.9.01.

\bibitem{redmon2015real}
J.~Redmon and A.~Angelova, ``Real-time grasp detection using convolutional
  neural networks,'' in \emph{2015 IEEE International Conference on Robotics
  and Automation (ICRA)}.\hskip 1em plus 0.5em minus 0.4em\relax IEEE, 2015,
  pp. 1316--1322.

\bibitem{karaoguz2019object}
H.~Karaoguz and P.~Jensfelt, ``Object detection approach for robot grasp
  detection,'' in \emph{2019 International Conference on Robotics and
  Automation (ICRA)}.\hskip 1em plus 0.5em minus 0.4em\relax IEEE, 2019, pp.
  4953--4959.

\bibitem{osawa2007unfolding}
F.~Osawa, H.~Seki, and Y.~Kamiya, ``Unfolding of massive laundry and
  classification types by dual manipulator,'' \emph{Journal of Advanced
  Computational Intelligence and Intelligent Informatics}, vol.~11, no.~5, pp.
  457--463, 2007.

\bibitem{bersch2011bimanual}
C.~Bersch, B.~Pitzer, and S.~Kammel, ``Bimanual robotic cloth manipulation for
  laundry folding,'' in \emph{2011 IEEE/RSJ International Conference on
  Intelligent Robots and Systems}.\hskip 1em plus 0.5em minus 0.4em\relax IEEE,
  2011, pp. 1413--1419.

\bibitem{jimenez2017visual}
P.~Jim{\'e}nez, ``Visual grasp point localization, classification and state
  recognition in robotic manipulation of cloth: An overview,'' \emph{Robotics
  and Autonomous Systems}, vol.~92, pp. 107--125, 2017.

\end{thebibliography}

\end{document}